\documentclass[sn-mathphys-num]{sn-jnl}

\usepackage{graphicx}%
\usepackage{longtable}
\usepackage{multirow}%
\usepackage{amsmath,amssymb,amsfonts}%
\usepackage{amsthm}%
\usepackage{mathrsfs}%
\usepackage[title]{appendix}%
\usepackage{xcolor}%
\usepackage{textcomp}%
\usepackage{manyfoot}%
\usepackage{booktabs}%
\usepackage{algorithm}%
\usepackage{algorithmicx}%
\usepackage{algpseudocode}%
\usepackage{listings}%
 \usepackage{float} 

\theoremstyle{thmstyleone}%
\theoremstyle{thmstyletwo}%

\theoremstyle{thmstylethree}%

\begin{document}

\title[Article Title]{\textbf{CataLM}: Empowering Catalyst Design Through Large Language Models}


\author[1]{\fnm{Ludi} \sur{Wang}}\email{wld@cnic.cn}
\equalcont{These authors contributed equally to this work.}

\author[1,3]{\fnm{Xueqing} \sur{Chen}}\email{xqchen@cnic.cn}
\equalcont{These authors contributed equally to this work.}
\author[1,3,4]{\fnm{Yi} \sur{Du}}\email{duyi@cnic.cn}
\author[1,3,4]{\fnm{Yuanchun} \sur{Zhou}}\email{zyc@cnic.cn}
\author*[2]{\fnm{Yang} \sur{Gao}}\email{gaoyang@nanoctr.cn}
\author*[1,3]{\fnm{Wenjuan} \sur{Cui}}\email{wenjuancui@cnic.cn}

\affil[1]{Laboratory of Big Data Knowledge, Computer Network Information Center, Chinese Academy of Sciences, Beijing, 100083, China}
\affil[2]{CAS Key Laboratory of Nanosystem and Hierarchical Fabrication, National Center for Nanoscience and Technology (NCNST), Beijing 100190, China}
\affil[3]{University of Chinese Academy of Sciences, Beijing, 100049, China}
\affil[4]{Hangzhou Institute for Advanced Study, UCAS, Hangzhou, 310000, China}


\abstract{The field of catalysis holds paramount importance in shaping the trajectory of sustainable development, prompting intensive research efforts to leverage artificial intelligence (AI) in catalyst design. Presently, the fine-tuning of open-source large language models (LLMs) has yielded significant breakthroughs across various domains such as biology and healthcare. Drawing inspiration from these advancements, we introduce \textbf{CataLM} (\textbf{Cata}lytic \textbf{L}anguage \textbf{M}odel), a large language model tailored to the domain of electrocatalytic materials. Our findings demonstrate that \textbf{CataLM} exhibits remarkable potential for facilitating human-AI collaboration in catalyst knowledge exploration and design. To the best of our knowledge, \textbf{CataLM} stands as the pioneering LLM dedicated to the catalyst domain, offering novel avenues for catalyst discovery and development.}




\maketitle

\section{Introduction}\label{sec1}
The field of catalysis is crucial to the future of sustainable development. Innovative catalysts can generate clean fuels, reduce the impact of global warming and provide solutions to environmental pollution\cite{de2019would, seh2017combining}. Theoretical calculations and simulations can accelerate catalyst screening through activity descriptors that link structure to catalyst activity\cite{norskov2009towards, suntivich2011perovskite, liu2020progress}. However, numerous variables exist in the synthesis, composition, structure, and performance of electrocatalysts, with much of this critical knowledge often elusive within scientific literature. This poses challenges in elucidating the intricate correlations from limited experimental data. Artificial intelligence can be used to extract, analyze and understand key information embedded in the vast scientific literature on catalysis that can be dedicated to predicting new catalysts. Natural language processing techniques and generative language models enable the extraction and analysis of textual information from scientific literature and the generation of domain-relevant on-demand text, which has shown potential in recent biological and thermoelectric research. However, language models in the field of catalysis are sparse and limited in scale, which restricts their use in empowering knowledge extraction in catalytic materials research and further enabling the discovery of new catalysts.

Pre-trained models have demonstrated their powerful capabilities in Natural Language Processing (NLP). There are two main types of pre-trained models: (1) BERT-like models\cite{devlin2018bert, liu2019roberta, clark2020electra}, which are mainly used for language comprehension tasks, and (2) GPT-like models\cite{radford2018improving, radford2019language, brown2020language}, which are mainly used for language generation tasks. Currently, large-scale language models (LLMs) such as GPT-4.0\cite{openai2023gpt} have laid a solid foundation for various applications. Although current large language models are effective in general domains, they often fail to meet the needs of catalytic scientists. Much of this inadequacy is attributed to the lack of reliable knowledge about catalysts, as relevant catalyst structural features and performance analyses are rarely present in commonly used pre-trained text corpora, such as C4\cite{raffel2020exploring} and the Pile\cite{gao2020pile}. Furthermore, the best performing large language models like ChatGPT are only served through APIs, which creates a barrier to research and progress in external domains. Fine-tuning open-source large language models is an effective way to meet domain-specific needs.

Currently, fine-tuning open-source large language models have reached considerable success in fields such as biology, healthcare, and finance. In biology, a domain-specific pre-trained Transformer language model, BiOGPT\cite{luo2022biogpt}, has been developed for biomedical text generation and mining. The model can be optimised and enhanced for performance in tasks such as biological named entity tasks and protein molecular design. In healthcare, models like HuatuoGPT\cite{zhang2023huatuogpt} and DoctorGLM\cite{xiong2023doctorglm} have been developed to address healthcare challenges, which exhibit a high degree of expertise and provide valuable insights into the healthcare domain. In recent years, researchers have utilized existing databases such as Atomly\cite{xie2023lu}, OQMD\cite{saal2013materials}, MaterialsProject\cite{jain2013commentary} and others. They have successfully explored the complex relationship between material structure and properties\cite{liang2023universal}, addressing the challenges posed by the scarcity of material data by developing more accurate AI optimisation\cite{liu2022swarm} and training methods\cite{guo2022neural}. With the application of LLMs, materials science researchers have explored the use of these models to address challenges such as chemical reactions and the complex nature of structures. Examples include the MatSciBERT\cite{gupta2022matscibert} model for the task of materials named entity recognition. MatSciBERT uses a large amount of materials science literature to fine-tune the BERT model\cite{devlin2018bert}, demonstrating the ability to automatically extract information from the literature, perform data mining, and construct knowledge graphs. MatChat\cite{Chen_2023} optimises the LLaMA2-7B model using knowledge of inorganic materials science literature and presents a viable solution for predicting chemical synthesis pathways of inorganic materials, opening up new possibilities for the use of language models in materials science. To the best of our knowledge, there has been no reported utilization of large language models in catalyst science so far.

In this work, we provide \textbf{CataLM}, a large language model aligned with knowledge in the field of electrocatalytic materials. This large language model takes advantage of the pre-trained Vicuna-13B model, and is trained on domain literature and data annotated by experts. With this extensive and diverse data, the original LLM is specialized with two phases: \texttt{Domain Pre-training}, where the model harvests the chemical knowledge from domain field literature, and \texttt{Instruction Tuning}, where the model further understands the requirements of downstream task with the annotation data. We use two tasks to validate \textbf{CataLM}, namely entity extraction task and control method recommendation task. In addition to using the constructed knowledge base for validation, we also invited domain experts to evaluate the answers of \textbf{CataLM} to verify its generalization ability. Results show that our large language model has potent potential for human-AI collaboration in catalyst knowledge search and design. To the best of our knowledge, \textbf{CataLM} is the first LLM that focus on the catalyst domain field, and we believe it can bring new possibilities for the preparation of new catalysts.

\section{Related Work}\label{sec2}
ChatGPT was selected as one of Nature's Top 10 Individuals of 2023, marking the unprecedented selection of a computer program—the first non-human entity in history—to receive such recognition. Nature states that this award aims to recognize the role of large language models (LLMs) in scientific development and progress. In the field of materials, numerous studies have utilized language models to address diverse tasks. Chen et.al provide the model MatChat \cite{Chen_2023}, for predicting inorganic material synthesis pathways. Xie et.al \cite{xie2023large} use FAIR database to fine-tune LLMs and design a downstream task named SII which aims to extract hierarchical, domain specific material and device information, such as composition, structure, preparation conditions, etc., from unstructured scientific texts. Zheng et.al \cite{doi:10.1021/jacs.3c05819} used prompt engineering to guide ChatGPT in the automation of text mining of metal–organic framework (MOF) synthesis conditions from diverse formats and styles of the scientific literature. InstructMol\cite{cao2023instructmol} adopts Vicuna 
to multiple chemical tasks with task-specific fine-tuning. Zheng and colleagues utilized prompt engineering to direct ChatGPT in automating text mining for the synthesis conditions of metal-organic frameworks\cite{zheng2023chatgpt}. However, previous works focus on the development of new materials instead of new catalyst designing. Considering the diversity of structural characteristics such as composition, crystal structure, and crystal plane of materials, potential catalysts are very abundant.Secondly, domain fine-tuning data sets which are consistent with downstream applications are crucial for the capability migration of LLMs, which is lacking in the field of catalyst design. This deficiency results in the model's lack of catalyst knowledge, making it challenging to achieve satisfactory parameters. 

To promote the creative utilization of large language models in catalysts science, this study utilizes a meticulously crafted database for question-answering to investigate their capabilities in the field of catalysts science. While building this model, we also refer to the successful experiences in the field of other science domains. For example, DeepGO-SE\cite{kulmanov2024protein} tries to predict GO functions from protein sequences using a pretrained large language model. MedPaLM2\cite{qian2024liver} and PMC-LLaMa\cite{wu2023pmc} attempt to tailor LLMs specifically for the fields of biology and medicine through fine-tuning with domain-specific instructions.

\section{CataLM}\label{sec3}
\begin{center}
\begin{figure}[h]
\includegraphics[width=1\linewidth]{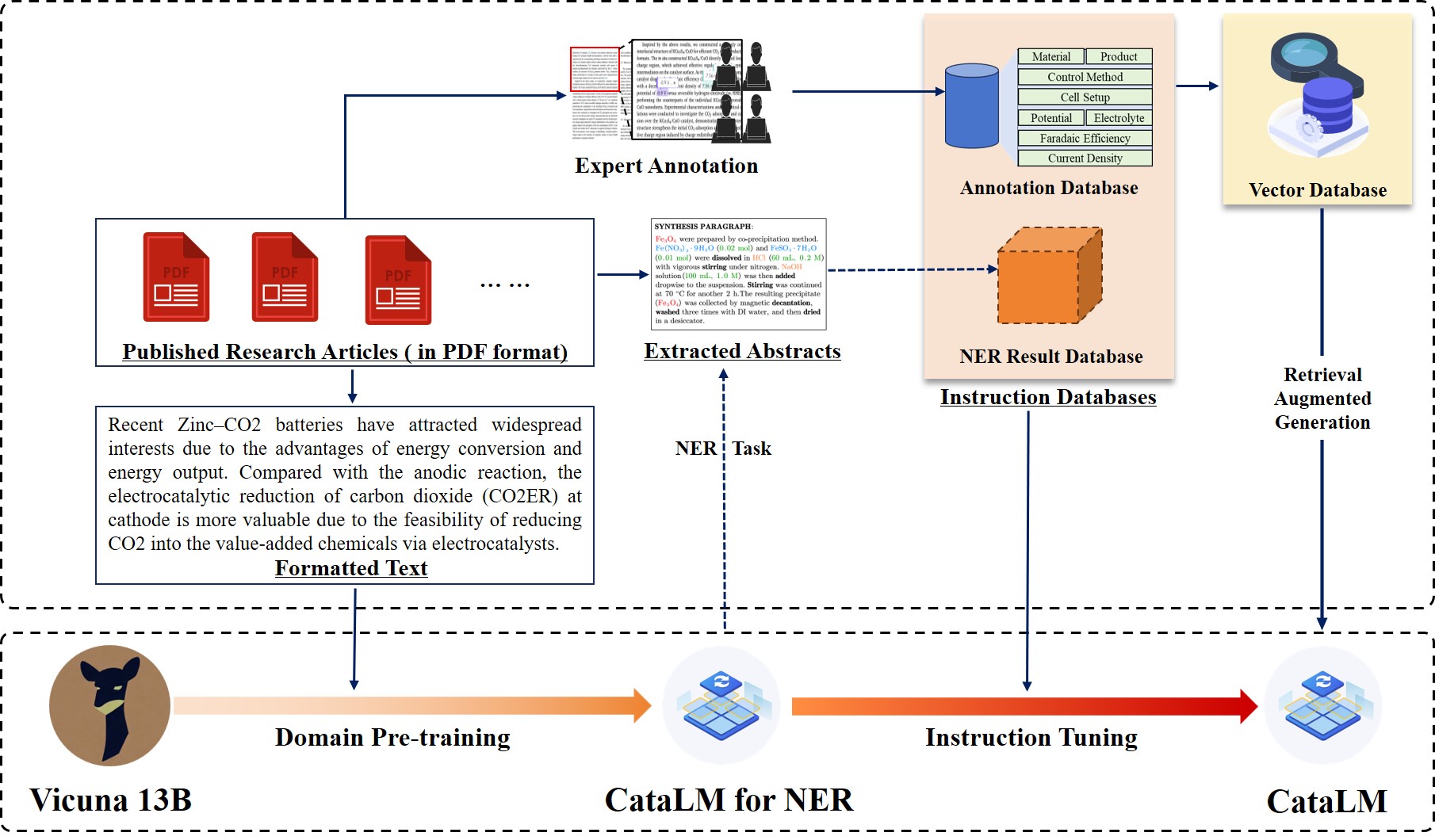}
\caption{\centering\label{figworkflow} The training pipeline of \textbf{CataLM}. The bottom part illustrates the primary training pipeline of CataLM, while the top part of the figure delineates the entire data preparation process for training. }
\end{figure}
\end{center}

As shown in Figure \ref{figworkflow}, the training of \textbf{CataLM} consists of two stages, which are Domain Pre-training and Instruction Tuning respectively. Due to the lack of open-source corpora for recommending catalyst control methods, we utilized expert annotated corpora, as well as the retrieval enhanced corpora generated by large language models for training during the instruction fine-tuning stage.
\subsection{Domain Pre-training}\label{subsec3.1}
In this work, the text corpus we used to further pre-train Vicuna-33b-v1.3, including the full text of open-access catalytic papers published in selected high-quality journals in the field of electrocatalytic science. We used Web of Science to find scientific literature on electrocatalytic CO2 reduction. Specifically, we exported the metadata of more than 22,000 articles from Web of Science using the keywords "CO2", "Reduction", and "Electro*" as subject indexes. Eventually, we used the full-text PDFs of 12,643 open-access papers to build the text corpus.

\textbf{\emph{PDF Parsing}}. We build an automatic PDF parsing toolkit based on the PyMuPDF library\cite{liu2018pymupdf}. Since the processed documents contain irrelevant tags, we developed a data cleaning method for parsing article tag strings into consistently formatted text paragraphs while retaining the same section and paragraph structure as the original paper. Finally, we use regular expressions and rule-based scripts to clean the data, removing the text obstructing reading, garbled, and impurity data.

\textbf{\emph{Vector Database}}. Despite the fact that Language Model Models (LLMs) are capable of responding to broad inquiries, they are limited in their ability to provide in-depth, precise, and timely information within specific vertical domain. To tackle this issue, we have employed vector databases to augment the reasoning capabilities of LLMs in vertical domain contexts. Vector databases can transform literature and data into vector representations through the process of embedding vectors. For the establishment of vector databases, Sci-BERT\cite{beltagy2019scibert} has been utilized as an embedding model.

The study involved retrieving titles and abstracts from a dataset containing 12,643 documents, manually annotating catalytic reaction processes by domain experts, and then merging and converting these textual elements into vector representations using Sci-BERT as an embedding framework. In the context of a catalytic domain-specific task such as Name Entity Recognition (NER), the embedding model operates by converting the user query into vector form. Relevant articles are then identified by vector distance calculations to facilitate the retrieval of accurate and relevant information.

\subsection{Instruction Tuning}\label{subsec3.2}
In order to align pre-trained models with domain user intent, we need to construct instruction tuning datasets. Currently available generic instruction tuning datasets such as Alpaca-GPT4\cite{wang2022self} and ToolBench\cite{qin2023toolllm} can only teach models to follow human instructions. For the specialized field of catalytic materials, we need to train models with knowledge-intensive data that can reflect domain knowledge. Considering the relatively small sample of data annotated by the experts, we use the large language model pre-trained in the previous section to expand it by automatically extracting abstracts from 12,643 documents. The entities were extracted based on an expert-constructed system of electrocatalytic reduction systems for literature content, including materials, conditioning methods, products, faradaic efficiency, cell setup, electrolyte, synthesis method, current density and voltage. The specific meanings and dataset formats of these entities can be found in the previous corpus construction work\cite{wang2023corpus, chen2024large}.

Firstly, we invite experts in the field of catalysis to perform manual annotation using a well-developed annotation tool, Autodive\cite{du2023autodive}. This tool allows annotators to access material literature through a web browser, view sentences for annotation, and interact with predefined entity types and descriptions. Annotators have the flexibility to include new entities, rearrange existing ones, or make edits in a separate view. We end up with a standard corpus\cite{wang2023corpus} in the field of electrocatalytic CO2 reduction containing 6,985 entities, with each record containing the entity extracted from the paper, its corresponding label, and the context sentence in which the entity is located. The standard corpus is provided as a file in CSV format, and the details are shown in Table 1.

\begin{table*}[ht]
\center
\caption{\centering\label{tab:3}The summary of the standard corpus}
\begin{tabular}{| p{6cm}|l|} \hline
Entity Type & Benchmark Corpus \\ \hline
Material&1,092\\ \hline
Control method&1,086\\ \hline
Product (including the second and third product)&1,340\\ \hline
Faradaic efficiency (including the Faradaic efficiency of second and third product)&1,135\\ \hline
Cell setup&435\\ \hline
Electrolyte&475\\ \hline
Synthesis method&228\\ \hline
Current density&393\\ \hline
Voltage&801\\ \hline
Total&6,985\\ \hline
\end{tabular}
\end{table*}

Next, we use the pre-trained large language model based on vector database augmentation from the previous section to perform automatic extraction of literature abstracts in the field of catalysis, which extracts a total of 30283 entities. It is important to highlight that the synthetic method of expert annotation in the dataset is an unstructured text paragraph description. We used a multi-model algorithm combining pattern recognition and neural networks to convert it into a structured synthetic pathway\cite{chen2024large} containing information about the prepared and target materials, synthetic operations and operating conditions. This structuring of information enhances the interpretability of domain knowledge by the expansive models.

The final dataset used to fine-tune the model in this paper consist of the according electrocatalytic CO2 reduction processes extracted from 12,643 papers. After rigorous filtering, de-duplication and cleaning, we obtained a training set consisting of 13,432 highly reliable catalytic process descriptions. Next, this dataset is further pre-processed and integrated into an instruction question-answering format. For example, for a certain catalytic reaction, using the entities provided in the dataset, we can reconstruct it as a recommendation task for catalyst preparation for a given product. As shown in Figure 2, the prompt involves a specific catalyst material query for a given product, and the answer provides the recommended material and its preparation method.

\begin{figure}[H]
\centering
\includegraphics[width=0.8\linewidth]{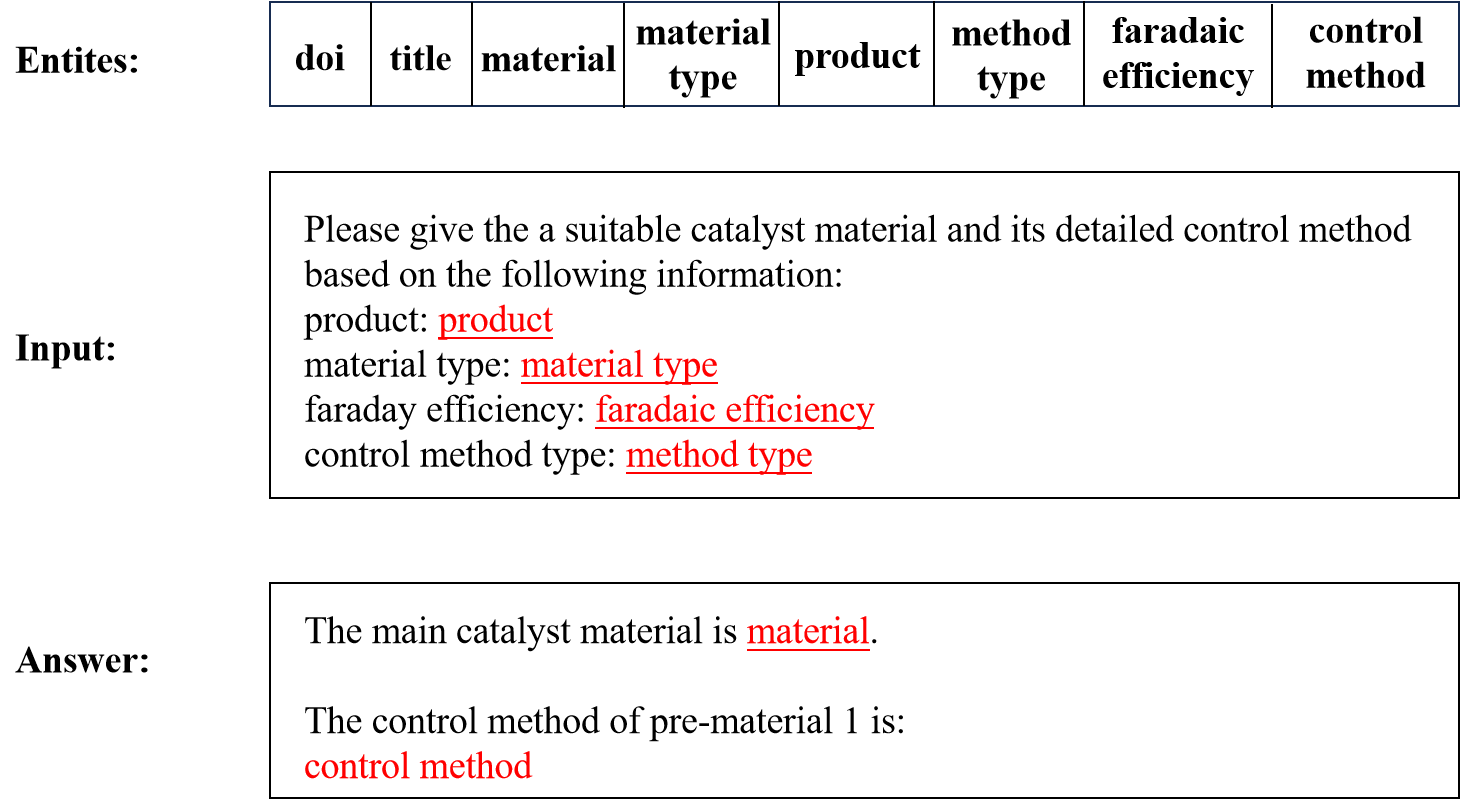}
\caption{\centering\label{figprompt1} Catalytic Material Recommended Scenario's Command Format.}
\end{figure}

\subsection{Training process}\label{subsec3.3}
The parameters of the model fine-tuning process are list in Table \ref{tab:parameters}. We use NVIDIA A100 GPUs for training, and techniques such as low-rank adaptation\cite{hu2021lora} is adopted to save storage memory and accelerate the process. Low Rank Adaptation of Large Language Models, also known as LoRA, is a technology developed by Microsoft researchers to address fine-tuning of large language models. The approach of LoRA is to freeze the pre-trained model weight parameters, and then inject trainable layers into each Transformer block. Since there is no need to recalculate the gradient of the model weight parameters, it greatly reduces the computational workload that needs to be trained. Research has found that the fine-tuning quality of LoRA is comparable to that of full model fine-tuning, thus we chose this method in the training process of \textbf{CataLM}.
\begin{table}[h]
\caption{\centering\label{tab:parameters} Parameter set.}
\renewcommand\arraystretch{1.5} 
\begin{tabular}{|p{3cm}|p{3cm}|} \hline
Parameter & Value \\\hline
batch size & 10\\ \hline
learning rate & 3*10$^{-4}$\\ \hline
lora r & 8\\ \hline
lora alpha & 32\\ \hline
lora dropout & 0.1\\ \hline
\end{tabular}
\end{table}
\section{Evaluation}\label{sec4}
\subsection{Named Entity Recognition Task}\label{subsubsec4.1}
The first task is named entity recognition, which aims to extract entity from the abstract of given literature. In this task, we use a dataset of 12,643 abstract from electrocatalytic scientific literature (the full text of these literature also be used in the fine-tuning of \textbf{CataLM}) for named entity recognition. We extracted eight types of entity labels, including material, control method, product, faradaic efficiency, cell setup, electrolyte, current density, and voltage. When performing entity recognition, the user first inputs the text to be extracted, and the embedding model transforms it into vectors. Then the similar articles will be obtained by calculating the vector distance, and will be used to generate precise and pertinent information, which be shown in Figure \ref{figprompt1}. The prompt will be fed into the fine-tuned LLM for entity recognition. 
\begin{figure}[h]
\centering
\includegraphics[width=0.8\linewidth]{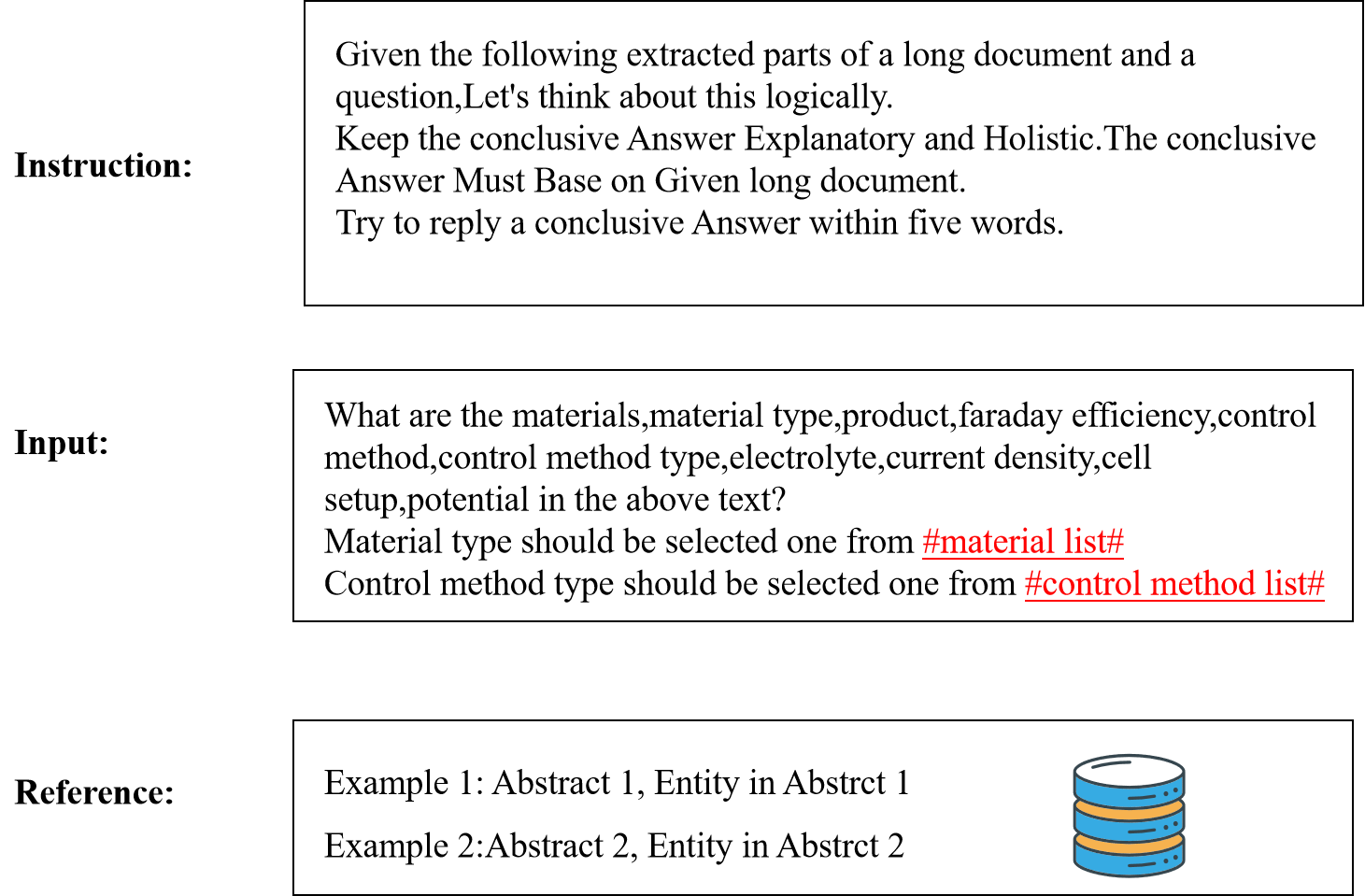}
\caption{\centering\label{figprompt1} Prompt in the named entity recognition task.}
\end{figure}

 For the evaluation and validation of the the entity extraction capability of \textbf{CataLM}, we randomly select 160 entries and validate the LLM's answers for them by experts, and ensure that each category has 20 test data. The evaluation result is shown in Table \ref{tab:evaluation}. The Count represents the total amount of samples from different categories, the Correct represents the number of correctly identified entities, and the Existence represents the number of entities of this type that do exist in the text input to the large language model. It is worth mentioning that if there is indeed no corresponding entity in the text input to the large language model, the situation where the large language model answers empty should also be considered as correct recognition. Therefore, we use Modified Correct to remove the above influence. Ultimately, we utilize Modified Correct and Count to calculate the evaluation of LLMs, which is Modified Accuracy.
 
\begin{table}[h]
\caption{\centering\label{tab:evaluation} The evaluation of entity recognition of \textbf{CataLM}.}
\renewcommand\arraystretch{1.5}
\setlength{\tabcolsep}{3pt}
\begin{tabular}{|l|l|l|l|l|l|} \hline
Entity & Count & Correct & Existence & Modified Correct & Modified Accuracy \\ \hline
MATERIAL & 20 &17&17&15&75\%\\ \hline
CONTROL METHOD&20&19&19&13&65\%\\ \hline
PRODUCT&20&17&17&17&85\% \\ \hline
FARADAIC EFFICIENCY&20&11&11&18&90\% \\ \hline
ELECTROLYTE&20&10&10&10&50\% \\ \hline
POTENTIAL&20&7&7&16&80\% \\ \hline
CURRENT DENSITY&20&7&7&12&60\% \\ \hline
CELL SETUP&20&6&6&9&45\% \\ \hline
\hline
OVERALL&160&85&94&110&68.75\% \\ \hline
\end{tabular}
\end{table}
From the results, we can see that \textbf{CataLM} performs better in entity extraction for numerical classes (faraday efficiency, potential, etc.), but performs poorly in entity extraction for descriptive classes. This may be due to the objectivity of data entities, which reduces the possibility of hallucinations in large language models.

We also conducted ablation experiments in this paper. We decomposed the model into two modules, namely the model Fine-tuning module and the Retrieval-Augmented Generation (RAG) module, and they were combined in pairs to form four possibilities. From Table \ref{tab:ablation experiment}, it can be seen that our method (i.e. Fine-tuned LLM + Few shot) performs the best. We can also see that both the fine-tuned module and the RAG module contribute to the improvement of model extraction accuracy.
\begin{table}[h]
\caption{\centering\label{tab:ablation experiment} Results of ablation experiment}
\renewcommand\arraystretch{1.5}
\setlength{\tabcolsep}{3pt}
\begin{tabular}{|l|l|l|l|} \hline
Model & Correct & Modified Correct & Modified Accuracy \\ \hline
Original LLM + Zero shot & 27 & 59 & 36.88\%\\ \hline
Original LLM + Few shot & 37 & 66 & 41.25\%\\ \hline
Fine-tuned LLM + Zero shot & 49 & 85 & 53.12\% \\ \hline
\textbf{Our method} & \textbf{85} & \textbf{110} & \textbf{68.75\% }\\ \hline

\end{tabular}
\end{table}
\subsection{Control Method Recommendation Task}\label{subsubsec4.2}
With the continuous development of big data technology, basic scientific research has shifted from the traditional "random trial and error" to the "data-driven AI" scientific model. Domain experts have also begun to attempt to use large language models to promote scientific innovation, such as literature understanding and summarization, experimental scheme generation, as well as unmanned experimental systems and scientific data sharing platforms, in order to improve scientific research efficiency and promote scientific progress and development. \textbf{CataLM} focuses on the scientific problems in the Catalyst Control field, and tries to assist scientists in catalyst design.

\begin{figure}[h]
\centering
\includegraphics[width=1\linewidth]{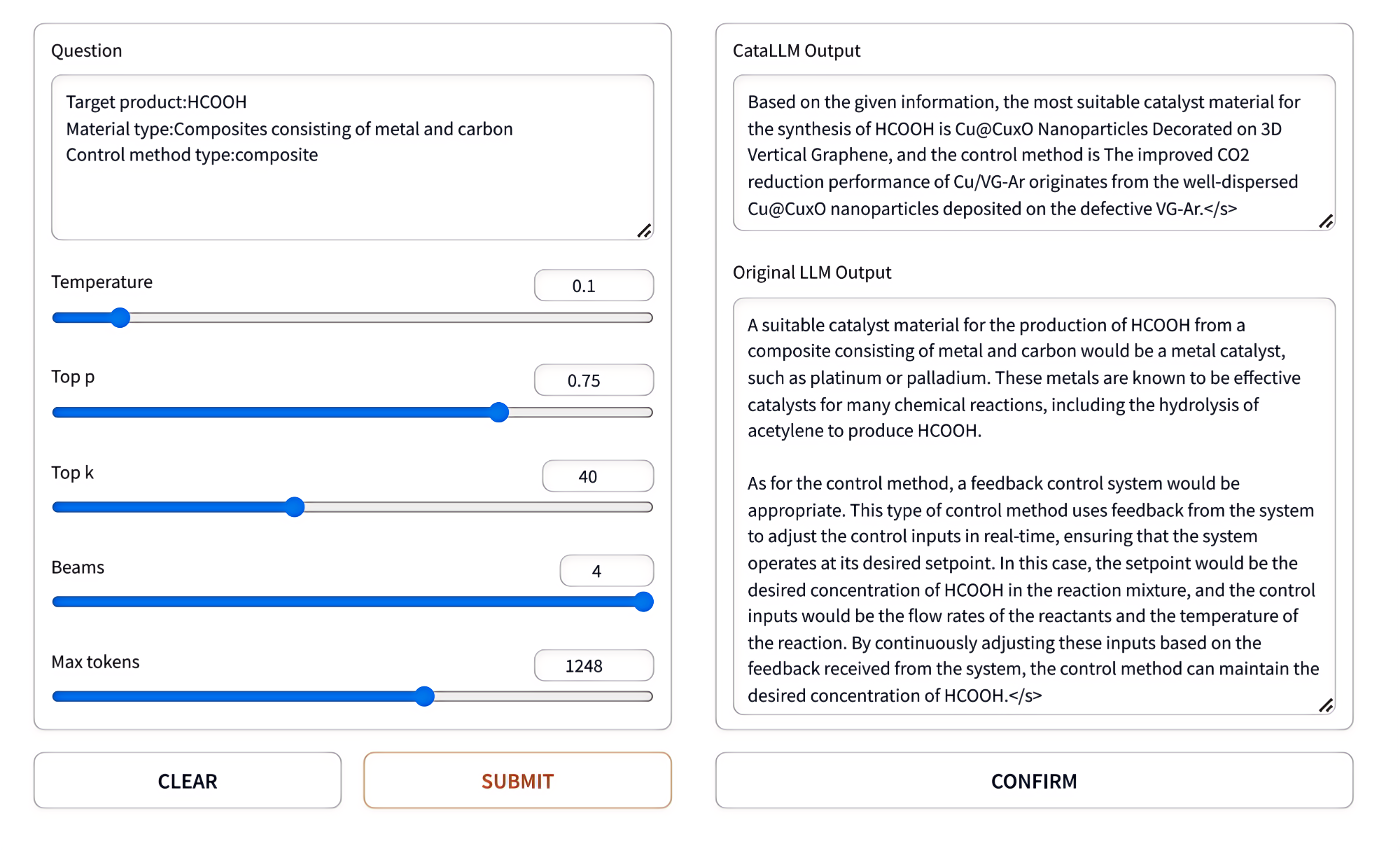}
\caption{\centering\label{answer} Answer from \textbf{CataLM} and original LLM.}
\end{figure}

However, how to evaluate the effectiveness of recommended catalyst control methods is a challenge faced by \textbf{CataLM}. In this paper, we invite domain experts to evaluate and analyze the recommendation methods generated by \textbf{CataLM} and the original LLM. Several representative results are listed in Table \ref{tab:answer}. The original LLM typically offers vague material selections and often lacks domain-specific expertise. Additionally, its understanding of control methods usually involves mere restatements of definitions, sometimes leading to significant misconceptions. In contrast, \textbf{CataLM} consistently integrates domain knowledge to provide more specific and accurate material choices. Its responses are logically coherent, with reasonable explanations of the mechanisms behind the selected materials. As illustrated in Figure \ref{answer}, \textbf{CataLM} demonstrates its ability to accurately interpret control methods and offer reasonable suggestions, such as the composite system of Cu and carbon, along with clear explanations for the choice of carbon materials. In contrast, the original LLM's misunderstanding of the control method is evident in its incorrect suggestion of Pt, which is not a catalyst for CO$_2$ reduction.

\begin{longtable}{|p{3cm}|p{4.5cm}|p{4.5cm}|}
\caption{\centering\label{tab:answer} The input and answer provided by orginal LLM and \textbf{CataLM}.}\\
\hline
Input & Original LLM & \textbf{CataLM} \\ \hline
\endhead
C2H5OH, Single metal, structure control & The suitable catalyst material for the production of C2H5OH (ethanol) would be a noble metal such as platinum or palladium. These metals are known for their ability to catalyze a wide range of chemical reactions and are often used in the production of ethanol through the hydrolysis of ethylene. As for the control method type, structure control would be the most appropriate. & Based on the given information, the most suitable catalyst material for producing C2H5OH is Palladium (Pd). \textbf{The control method that should be used is creating high-facets of wrinkled Pd surrounded by PdO mesh patterns.} \\\hline
C2H4, Metal oxide, structure control & A suitable catalyst material for the production of C2H4 (ethylene) would be a metal oxide such as iron oxide or copper oxide. These materials are known to be effective in the production of ethylene through the cracking of hydrocarbons.As for the control method, structure control would be an appropriate method to use. This type of control method involves manipulating the structure of the catalyst material in order to optimize its performance. & Based on the information provided, the most suitable catalyst material for producing C2H4 is OxygenCu (OCu) combination. \textbf{The control method that should be used is anodized copper (AN-Cu) Cu(OH)(2) catalysts by a simple electrochemical synthesis method.}  \\\hline
HCOOH, Composites consisting of metal and carbon, composite & Based on the given information, a suitable catalyst material for the production of HCOOH would be a composite consisting of metal and carbon. One example of such a catalyst is platinum-on-carbon (Pt/C).As for the control method, a suitable method would be to use a feedback control system. This type of control system uses a sensor to measure the concentration of HCOOH in the reaction mixture, and adjusts the reaction conditions in real-time to maintain a constant concentration of HCOOH. & Based on the given information, the most suitable catalyst material for the synthesis of HCOOH would be a composite of Copper nanoparticles embedded in Nitrogen-doped graphene oxide \textbf{(Cu/N-GO)}. The control method used in this composite would be the interaction between the copper nanoparticles and the nitrogen-doped graphene oxide support. \textbf{This interaction is expected to enhance the electrocatalytic reduction of CO2 to HCOOH, due to the synergistic effects between the copper nanoparticles and the nitrogen-doped}. \\\hline

CO, Alloys/composites of two or more metals, alloy & A suitable catalyst material for producing CO (carbon monoxide) from an alloy or composite of two or more metals would be a metal oxide such as iron oxide or cobalt oxide. As for the control method, a suitable option would be to use a feedback control system. & Based on the given information, the most suitable catalyst material for the target product CO is \textbf{Gold-Copper alloy}. The control method for this catalyst material is also \textbf{Alloy}. \\\hline
\end{longtable}

\section{Conclusion}\label{sec4}
In this paper, we introduce \textbf{CataLM}, a effective attempt towards catalyst design leveraging the capabilities of large language models. By undergoing domain pre-training and instruction tuning, our large language model has exhibited robust comprehension and reasoning skills in catalyst knowledge and patterns, achieving advanced performance in application tasks like knowledge extraction and recommendation of control methods. We have open sourced the \textbf{CataLM} model and fine-tuning data to facilitate further expansion and development by interested researchers, which is available at \hyperlink{https://github.com/kg4sci/CataLM.git}{https://github.com/kg4sci/CataLM}. The result of NER task is available at Science Data Bank (ScienceDB), which is a public, general-purpose data repository aiming to provide data services for researchers, research projects/teams, journals, institutions, universities, etc, the link is \hyperlink{https://www.scidb.cn/en/detail?dataSetId=3f6204bc48704fac9b64b8e95a904e02}{https://www.scidb.cn/en/detail?dataSetId=3f6204bc48704fac9b64b8e95a904e02}\cite{3f6204bc48704fac9b64b8e95a904e02}.

In the future, while continuously enhancing the field understanding ability of \textbf{CataLM}, we will also design and develop an auxiliary platform for field researchers based on it, in order to improve the efficiency of catalyst design work in practical applications. We believe that large language models will bring new and infinite possibilities to basic scientific research.


\section{Competing Interests}
The authors declare no competing interests.


\end{document}